\documentstyle[acra,algpseudocode,algorithm,amssymb,amsmath,array,float,graphicx,mathtools,url]{article}

\title{RANSAC: Identification of Higher-Order Geometric Features and Applications in Humanoid Robot Soccer}
\author{Madison Flannery$^1$, Shannon Fenn$^1$ and David Budden$^{2,3}$ \\$^1$ The University of Newcastle, Callaghan, NSW 2308, Australia.\\$^2$ National ICT Australia (NICTA), Victoria Research Lab\\$^3$ The University of Melbourne, Parkville, VIC 3010, Australia.\\david.budden@nicta.com.au}

\begin{document}

\maketitle

\begin{abstract}
The ability for an autonomous agent to self-localise is directly proportional to the accuracy and precision with which it can perceive salient features within its local environment. The identification of such features by recognising geometric profile allows robustness against lighting variations, which is necessary in most industrial robotics applications. This paper details a framework by which the random sample consensus (RANSAC) algorithm, often applied to parameter fitting in linear models, can be extended to identify higher-order geometric features. Goalpost identification within humanoid robot soccer is investigated as an application, with the developed system yielding an order-of-magnitude improvement in classification performance relative to a traditional histogramming methodology.

\end{abstract}

\section{Introduction}
\label{sec:introduction}

The problem of developing a team of humanoid robots capable of defeating the FIFA World Cup champion team, coined ``The Millennium Challenge", has been a milestone that has driven research in the fields of artificial intelligence, robotics and computer vision for over a decade~\cite{kitano1998robocup}. Critical among these challenges is self-localisation, described as the task of determining the coordinate transformation between the agent's local coordinate system and the environment's global coordinate system~\cite{thrun2005probabilistic}. Knowledge of this transformation allows the agent to consider global features with reference to its own coordinate frame, facilitating navigation and execution of complex actions.


Knowing the position and orientation of an agent is both sufficient and necessary for determining the local-to-global coordinate transformation. In a traditional robotics scenario, the agent employs physical sensors (such as cameras or range-finders) and computer vision algorithms to infer its relative position and orientation from salient landmark features. Previous research has demonstrated incremental improvement in self-localisation performance is possible by including domain-specific knowledge at different levels:

\begin{itemize}
\item Intelligent camera actuation policies with motivated reinforcement learning~\cite{fountain2013motivated}.
\item Refined probabilistic filters that incorporate constraints on observational noise models~\cite{budden2013particle}.
\end{itemize}

\noindent{Despite reduction in self-localisation error demonstrated in these two papers (11\% and 38\% respectively), the most simple and intuitive means of improving localisation performance remains the accurate identification of salient landmark features. Until recent years, such features in RoboCup have exhibited unique colour-coding, and therefore traditional computer vision research in this domain has focused on colour-based feature extraction methodologies~\cite{budden2012novel}. These algorithms rely on a dimensionality reduction from the original colour space to a small set of class labels, with this mapping generated offline and stored in a static look-up table data structure due to limited computational resources~\cite{budden2013colour}.

\begin{figure}[H]
\begin{center}
\begin{tabular}{>{\centering}p{3.8cm}<{\centering}  >{\centering}p{3.8cm}<{\centering}}
\includegraphics[width=3.8cm]{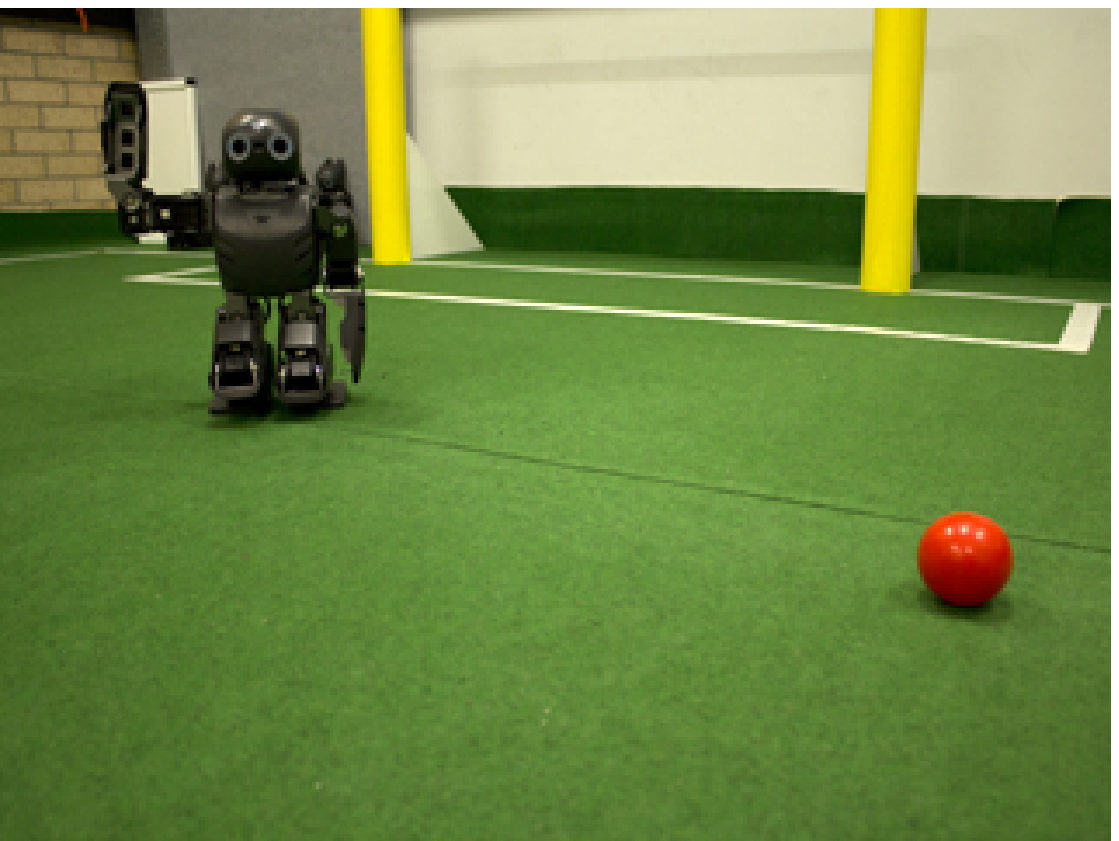}  & \includegraphics[width=3.8cm]{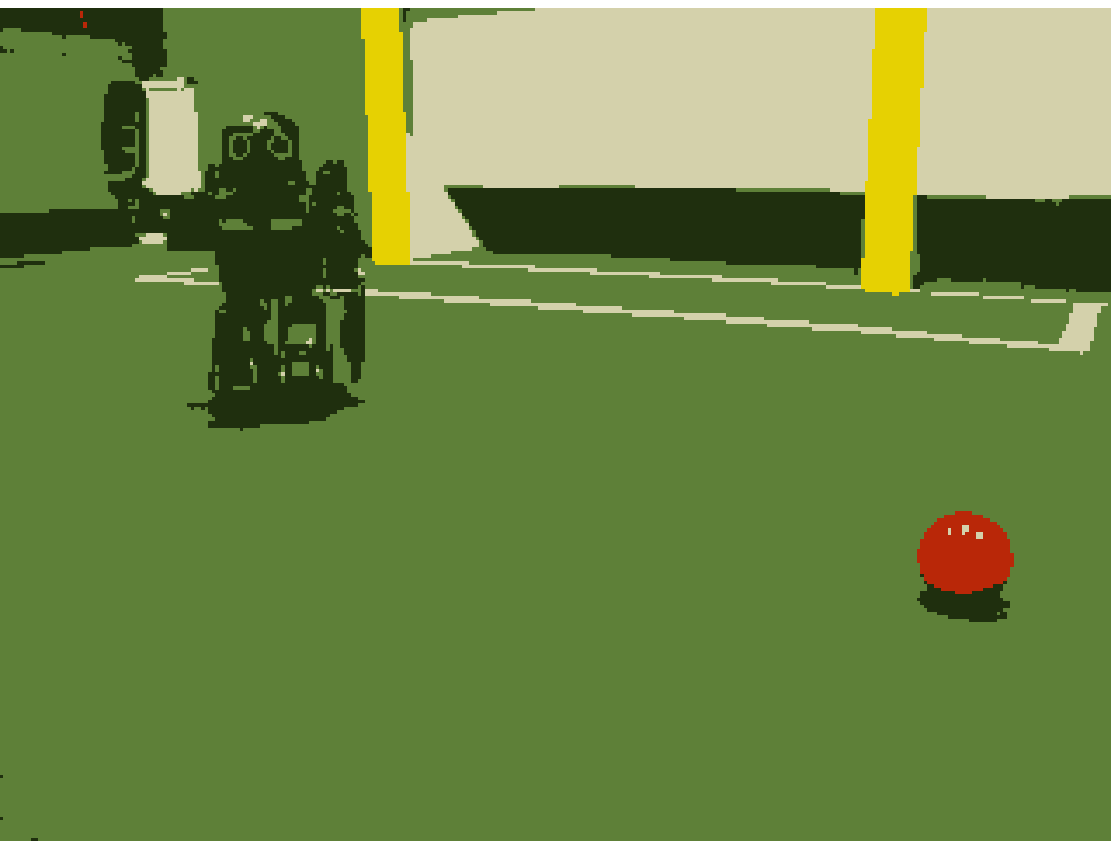}
\end{tabular}
\end{center}
\caption{Example of a raw RoboCup image (left) and the equivalent colour-segmented image (right). This segmentation was generated by the unsupervised framework described in [Budden and Mendes, 2013].
\label{fig:segmentation}}
\end{figure}

Although systems have been recently developed to allow the rapid, unsupervised generation of optimised colour look-up tables~\cite{budden2013salient}, their application are limited to domains exhibiting well defined coloured features (as demonstrated in Figure~\ref{fig:segmentation}), as well as spatiotemporal lighting invariance. For generalised applications of robots in industry (as well as present-day RoboCup), it is critical to develop feature extraction algorithms that are equally capable of handling images preprocessed by other image qualities (such as intensity, gradient or texture, or the output of convolutional filtering for edge detection).

This paper details a framework by which the random sample consensus (RANSAC) algorithm~\cite{fischler1981random} can be implemented to identify higher-order geometric features in a generalised robotics or object recognition scenario. RANSAC is commonly used in RoboCup for the extraction of simple environment features, with the University of New South Wales (rUNSWift~\cite{runswift2013}) and University of Newcastle (NUbots~\cite{nubots2013}) teams applying RANSAC-based methods for the respective identification of field border and field lines. Extending such techniques to identify more complex objects involves the following steps:

\begin{itemize}
\item Define the object of interest in terms of its 2-dimensional projection onto the image plane.
\item Define a set of rules that allow the decomposition of this projection into a set of geometric primitives (herein assumed to be lines for simplicity, although could be readily extended to include curves).
\item Identify primitives within the image:
    \begin{itemize}
    \item Generate a point cloud corresponding with potential feature edges within the image (using convolutional filtering, colour segmentation or any other applicable edge detection method).
    \item Apply RANSAC to fit (linear) model parameters to the points.
    \end{itemize}
\item Determine whether the identified primitives fulfil the decomposition rules.
\end{itemize}

Using goalpost identification in RoboCup soccer as an example, the remainder of this paper demonstrates a concrete implementation of this abstract framework: the 2-dimensional projections of goalposts are approximated as rectangular (decomposed into pairs of approximately parallel lines), with candidate points defined as the intersection of the boundaries of colour-segmented regions with a set of equidistant horizontal and vertical ``scan-lines". The performance of the algorithm is evaluated against a traditional 1-dimensional histogramming method, in terms of both explicit feature extraction accuracy and self-localisation performance.




\section{Colour Segmentation}
\label{sec:coloursegmentation}

As described in Section~\ref{sec:introduction}, the proposed RANSAC-based framework for feature extraction requires the generation of a point cloud corresponding with possible object edges (``candidate points"). Detection of goalposts in RoboCup soccer was chosen as a concrete implementation of this abstract framework, with an algorithm developed for the NUbots' team of DARwIn-OP humanoid robots~\cite{nubots2013}. As the NUbots' vision system is predominantly colour-based, these candidate points are defined as the intersection of the boundaries of colour-segmented regions with a set of equidistant horizontal and vertical ``scan-lines". The remainder of this section describes the specific implementation of this process for a RoboCup soccer environment, where lighting is relatively consistent and salient features exhibit unique and well-defined colour. For generalised applications of robots in industry, other methods of candidate point generation may be more appropriate (such as convolutional filtering for edge detection~\cite{marr1980theory}).

In computer vision, a mapping from an arbitrary 3-component colour space $C$ to a set of colours $M$ assigns a class label $m_{i} \in M$ to every point $c_{j} \in C$~\cite{budden2013colour}. If each channel is represented by an $n$-bit value and $k = |M|$ represents the number of defined class labels, then

\begin{equation*}
 C \rightarrow M,
\end{equation*}

\noindent{where}
\begin{equation*}
 C = \left\{0, 1, \dots, 2^{n}-1\right\}^{3} \; \text{and} \;
 M = \left\{m_{0}, m_{1}, \dots, m_{k-1}\right\}.
\end{equation*}
Where computational resources are limited, the colour segmentation process is performed off-line, with the resultant mapping represented in the form of a $2^{n} \times 2^{n} \times 2^{n}$ look-up table (LUT). Recent developments allow for the rapid generation of optimised LUTs (such as the one demonstrated in Figure~\ref{fig:lut}) using unsupervised machine learning techniques~\cite{budden2013salient}.

\begin{figure}[H]
\begin{center}
\includegraphics[width=5.0cm]{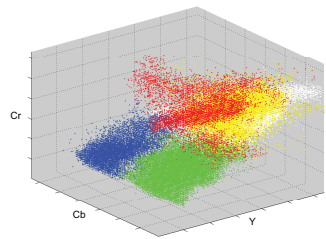}
\end{center}
\caption{A colour segmentation look-up table, mapping raw pixel values (in $YC_bC_r$ space) to colour class label.
\label{fig:lut}}
\end{figure}

Given an image captured by the DARwIn-OP robot and segmented using a colour look-up table, the NUbots' vision system follows the following steps to generate candidate points:

\begin{enumerate}
\item \textbf{Generate scan-lines:} As current processor limitations do not allow complex operations over every pixel of a 2-megapixel image without significant frame-rate reduction, only the pixels along a set of vertical and horizontal ``scan-lines" are considered for generating candidate points. This method is preferred over reduced camera resolution, as it provides a similar performance increase (i.e. the same number of pixels are considered) whilst still allowing for small, high resolution portions of the image to be processed to resolve finer detail. Scan lines may be either equidistant on the image plane (as per the NUbots's implementation), or spaced in such as way as to be equidistant on the field plane (requires robot kinematics data).
\item \textbf{Determine field border:} Identifying the field border allows RoboCup domain-specific knowledge to be considered in reducing the area of the image required for processing. For example, a ball and field lines should only ever be found beneath the border. Starting at the top of the image, each pixel along each vertical scan-line is inspected until a threshold of consecutive green pixels is exceeded, at which time the uppermost green pixel's coordinates are added to a list of points. The field border is defined as the upper convex hull of these points, calculated using a modified implementation of Andrew's monotone chain algorithm~\cite{andrew1979another}.
\item \textbf{Generate colour segments:} To generate colour segments, each pixel along each scan line is considered. Wherever the colour class label of a pixel differs from that of the previous adjacent pixel (known as a \emph{colour transition}), a segment is generated. The information stored in each segment includes its start, centre and end $(x,y)$ image coordinates, the length of the segment (in pixels) and its colour class label.
\item \textbf{Determine candidates:} Colour segments are processed by position, colour and length to determine which points are true candidates for the object of interest. Once unsuitable segments have been rejected, candidate points are defined as the centre $(x,y)$ coordinates of those that remain.
\end{enumerate}

More information regarding the specific implementation of the NUbots' vision system is available in their team description paper (\url{http://www.davidbudden.com/media/NUbotsTeamDescription2013.pdf}) and internally referenced publications~\cite{nubots2013}.

\section{1-Dimensional Colour Histogramming for Goalpost Identification}

As RoboCup goalposts traditionally exhibit a dominant vertical profile, maintaining a histogram of colour transitions or segments is traditionally adopted as a simple method of goalpost detection. This section describes the NUbots' 1-dimensional implementation of this method, which provides a benchmark for evaluating performance of the presented RANSAC-based approach.

In RoboCup soccer, a 1-dimensional image histogram $H = \left\{b^{[1]},b^{[2]},\cdots,b^{[N]}\right\}$ may be defined as a set of $N$ bins $b^{[n]}$, where each bin maintains a count $c^{[n]}$ of the length of all relevant colour segments, whose centres (i.e. the ``candidate points" described in Section~\ref{sec:coloursegmentation}) $t_m=(x_m,y_m)$ fall within a vertical column defined by some range $r^{[n]}$:

\begin{equation*}
x_m \in r^{[n]} = \left[x^{[n]}_{start}, x^{[n]}_{end}\right).
\end{equation*}

\noindent{We further define the ``peak candidates" $\tilde{P}$ as the set of all bins containing a minimum number $\gamma$ of transitions:}

\begin{equation*}
\tilde{P} = \left\{b^{[n]}\in H : c^{[n]} \ge \gamma \right\}.
\end{equation*}

To apply this definition to the task of goalpost identification, each image frame is initially split into $N = 20$ bins of equal width

\begin{equation} \label{eqn:hist1}
r^{[n]} = \left[\frac{(n-1)\times w}{N}, \frac{nw}{N}\right), \indent{\forall n\in \left[1, N\right]}
\end{equation}

\noindent{where $N$ is the number of bins, $w$ is the pixel width of the image, and $(x_m, y_m)$ are the pixel coordinates of the centre of some colour segment $t_m$. It should be noted that each set of bins is considered to maintain a strict ordering; all bins have disjoint ranges, and adjacent bins share a common boundary point.

\begin{figure}[H]
\begin{center}
\includegraphics[width=6.0cm]{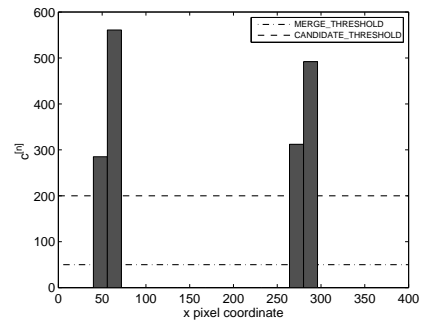}
\end{center}
\caption{Histogram demonstrating the count $c^{[n]}$ of all relevant colour segments, whose centres $t_m=(x_m,y_m)$ fall within a vertical column defined by some range $r^{[n]}$. Two potential goalpost candidates are evident. \label{fig:goalposthistogram}}
\end{figure}

Considering the assumption that all colour segments belonging to a goalpost exhibit the appropriate colour class label (yellow in the case of the RoboCup humanoid league), each yellow horizontal and vertical segment's length must then be added to the appropriate bin as specified in (\ref{eqn:hist1}). During this process, a simple standard deviation check is placed on the segment lengths; any segment with a length sufficiently larger than the standard deviation of all segment lengths is not added to the histogram (parameterised by some user-defined threshold value). This allows for the avoidance of large, unwanted segments being falsely identified as goal post candidates (for example, the top bar connecting the two posts). Figure~\ref{fig:goalposthistogram} demonstrates a histogram resulting from applying this process to a single image frame containing two goalposts.

Any adjacent peaks are merged for the purpose of goalpost candidate identification. Algorithm~\ref{alg:GroupPeaks} describes the implementation of this merge process on the peak candidates, $\tilde{P}$, by iterating through the histogram bins, locating adjacent peaks, and creating and adding the new peak to a set of final peaks, $P$.

\begin{algorithm}
\caption{Grouping peaks for 1-dimensional colour histogramming-based goalpost identification}
\label{alg:GroupPeaks}

\textbf{Inputs} \\
$b^{[n]}$: A set of bins.\\
$\tilde{P}$: A set of peak candidates.\\

\textbf{Outputs}\\
$P$: A set of intervals that represent adjacent peaks.
\line(1,0){245}
\begin{algorithmic}
\State $P \gets \varnothing$
\For{$i \in \left[1, N-1\right]$}
    \If{$b^{[i]} \in \tilde{P}$}
        \State $j \gets i$
        \While{$b^{[j+1]} \in \tilde{P}$}
            \State $j \gets j + 1$
        \EndWhile
        \State $b_{new} \gets [x^{[i]}_{start}, x^{[j]}_{end})$
        \State $P \leftarrow P + \lbrace b_{new}\rbrace$
    \EndIf
\EndFor\\
\end{algorithmic}
\end{algorithm}

To form the final goal post, a bounding box is generated by calculating the uppermost and lowermost $x$ and $y$ coordinates of all segments within each set of grouped peaks (with each corresponding with a single goal candidate). These values are considered to represent the four corners of the detected post.

\section{Extending RANSAC for Goalpost Identification}

Random sample consensus (RANSAC) is a stochastic algorithm for model fitting in datasets with a large proportion of outliers~\cite{fischler1981random}. It functions by initially fitting a random model, before separating the data points into two sets: a ``consensus" set and a ``non-consensus" set. After a fixed number of trials, the model with the largest consensus is kept. An extended version of the RANSAC algorithm, as applied to the fitting of multiple linear models to a set of data, is defined in Algorithm~\ref{alg:RANSAC}. The original algorithm requires two user-defined parameters: $d_{inliers}$, which defines the maximum distance a point can be from the fit line to be considered an inlier; and $k$, the number of fitting attempts to be made before keeping the best line. The extended version requires a further two parameters: $n$, the minimum number of points for a model, and $M_{max}$, the maximum number of models to find.

\begin{algorithm}[H]
\caption{RANSAC}
\label{alg:RANSAC}
\textbf{Inputs}\\
$\mathbf{P}$: A set of points in $\mathbb{R}$$^2$ \\
$d_{inlier}$: distance threshold for inliers\\
$k$: number of attempts per line\\
$n$: minimum number of points\\
$M_{max}$: maximum number of models\\
\\
\textbf{Output}\\
$\mathbf{L}$: A list of line segments.\\
\line(1,0){245}
\begin{algorithmic}
\Repeat
    \State $\mathbf{P_{best}} \gets \varnothing$
    \For{$i = 1 \to k$}
        \State Fit a line $l$ to two random points in $\mathbf{P}$
        \State $\mathbf{P_{inliers}} \gets \varnothing$
        \ForAll{$p\in \mathbf{P}$}
            \State $d \gets p_{\perp}l$
            \If{$d<d_{inlier}$}
                \State $\mathbf{P_{inliers}} \gets \mathbf{P_{inliers}} \cup \lbrace p \rbrace$
            \EndIf
        \EndFor
        \If{$\left|\mathbf{P_{inliers}}\right| > \left|\mathbf{P_{best}}\right|$}
            \State $\mathbf{P_{best}} \gets \mathbf{P_{inliers}}$
        \EndIf
    \EndFor
    \If{$\left|\mathbf{P_{inliers}}\right| \geq n$}
        \State Fit a line to $\mathbf{P_{best}}$ and add it to $\mathbf{L}$.
    \EndIf
    \State $\mathbf{P} \gets \mathbf{P} \setminus \mathbf{P_{best}}$
\Until{$\left|\mathbf{P}\right| < n$ \textbf{or} $\left|\mathbf{P_{best}}\right| = M_{max}$}
\\
\end{algorithmic}
\end{algorithm}

As RANSAC is a stochastic algorithm, a valid model will only be generated at each iteration with some probability $p$ (proportional to the parameter $k$). Specifically:
\begin{equation*}
p = 1 - (1 - q^2)^k,
\end{equation*}
\noindent{where $q$ is the probability of drawing a point fitting that particular model from the input data set:}
\begin{equation*}
q = \frac{\text{number of points in line}}{\text{total number of points}}.
\end{equation*}

\noindent{Note that $q$ varies between data sets, and cannot be explicitly used to determine an optimal $k$ value.}

The extended RANSAC implementation (as defined in Algorithm~\ref{alg:RANSAC}) was applied to generate multiple linear models from two sets of points: the set of points corresponding with the left-hand edge of a goalpost (indicated by green-yellow colour transitions), and the set corresponding with the right. The parameters $k = 50$, $n = 6$, $d_{inlier} = 5.0$ and $M_{max} = 3$ were empirically determined as robust against classification noise and other RoboCup environmental factors. The output from this process is two sets of lines, $\mathbf{L_s}$ and $\mathbf{L_e}$.

To refine the sets $\mathbf{L_s}$ and $\mathbf{L_e}$ into identified goalposts, a heuristic is implemented that determines line pairs $\left(l_1, l_2\right) : l_1 \in \mathbf{L_s},$ $l_2 \in \mathbf{L_e}$ that satisfy the following similarity conditions:
\begin{itemize}
\item $\theta < \epsilon _{ \theta }$, where $\theta$ is the acute angle formed between the lines, and $\epsilon_{\theta}$ is an upper bound controlling the permissiveness of this heuristic.
\item $|\ell_{1} - \ell_{2}| < \epsilon_{\ell}$, where $\ell_{1}$ and $\ell_{2}$ are the line segment lengths, and $\epsilon_{\ell}$ is an upper bound.
\item $\frac{d_{1} + d_{2}}{2} < \epsilon_{d}$, where $d_{1} = \min \lbrace |\boldsymbol{p_{11}} - \boldsymbol{p_{21}}| , |\boldsymbol{p_{11}} - \boldsymbol{p_{22})}| \rbrace$ and $d_{2} = \min \lbrace |\boldsymbol{p_{12}} - \boldsymbol{p_{21}}| , |\boldsymbol{p_{12}} - \boldsymbol{p_{22}}| \rbrace$ are the shortest distances between the endpoints of the first line segment ($\lbrace \boldsymbol{p}_{11} , \boldsymbol{p}_{12} \rbrace$) and the endpoints of the second ($\lbrace \boldsymbol{p_{21}} , \boldsymbol{p_{22}} \rbrace$). Averaging these distances provides a measure for the ``closeness'' of approximately parallel line segments (i.e. those satisfying the first condition). $\epsilon_{d}$ is an upper bound.
\item $|n_{1} - n_{2}| < \epsilon_{n}$, where $n_{1}$ and $n_{2}$ are the number of consensus points associated with the lines, and $\epsilon_{n}$ is an upper bound.
\end{itemize}

In order to minimise the number of parameters associated with this heuristic, each is defined in terms of a single permissiveness parameter, $\rho \in [0, 1]$, such that:
\begin{align*}
\epsilon_{\theta} &= \rho \cdot \frac{\pi}{2},\quad&\epsilon_{\ell} = \rho \cdot \max \lbrace \ell_{1} , \ell_{2} \rbrace, \\
\epsilon_{d} &= \rho \cdot \sqrt{w^{2} + h^{h}},\quad&\epsilon_{n} = \rho \cdot \max \lbrace n_{1} , n_{2} \rbrace.
\end{align*}

\noindent{Line pairs satisfying this heuristic are collected, and their endpoints, $\lbrace \boldsymbol{p}_{11} , \boldsymbol{p}_{12} , \boldsymbol{p}_{21} , \boldsymbol{p}_{22} \rbrace$, define the corners of the final identified goalpost.}

\section{Evaluation Metrics}
\label{sec:evalmethods}
To compare the performance of both histogramming and RANSAC-based goalpost identification methods, a series of 3600 images were captured using the DARwIn-OP robot's Logitech C905 camera~\cite{ha2011development}. The DARwIn-OP (shown in Figure~\ref{fig:darwin}) was placed at fixed distances, directly in line with the right goalpost on the $6 \times 4$ metre RoboCup field. These distances were chosen to correspond with salient RoboCup field positions; specifically 60, 150, 300, 350, 530 and 600 cm from the goalpost. 200 images were captured by the DARwIn-OP at each of these field positions, for coronal body tilts of 0, 10 and 20 degrees from vertical. Maintaining goalpost identification performance for small tilt angles is essential within RoboCup, due to the natural coronal oscillation of a walking bipedal robot.

Given the 3600 image frames captured by the DARwIn-OP~, two evaluation metrics were applied to evaluate the performance of each goalpost identification system: ``detection rate", which is simply the fraction of frames for which goalposts were correctly identified; and ``distance-by-width", which considers the calculated pixel width of a detected goalpost to estimate its distance from the robot. To quantify the accuracy of each identified post, the distance-by-width, $d_w$, from goalpost to robot is calculated as:
\begin{equation*}
d_w = \frac{w_{cm}}{w_{px}} \cdot \gamma_x, \indent{\gamma_x = \frac{w_{img}}{2\cdot\tan{\left(\frac{\theta_x}{2}\right)}}},
\end{equation*}

\noindent{where $w_{cm}$ is the actual width of the goalpost in cm and $w_{px}$ is the width of the goalpost located in the image in pixels. The parameter $\gamma_x$ defines the ``pixel angular width", which approximates the relationship between $\theta_x$ (the horizontal field of view of the camera) and $w_{img}$ (the width of the image in pixels). This calculation applies the knowledge that the vertical profile of the goal post is approximately parallel to a standing robot, removing the need for a pythagorean transformation and therefore knowledge of the height of the camera (which requires error-prone kinematics transformations).}

\begin{figure}[H]
\begin{center}
\includegraphics[width=2.4cm]{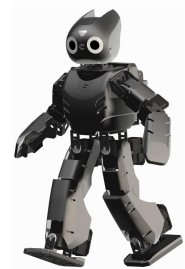}
\end{center}
\caption{The Robotis DARwIn-OP humanoid robot used as a platform for this research [Ha \emph{et al}., 2011].
\label{fig:darwin}}
\end{figure}

\section{Experimental Results}

As described in Section~\ref{sec:evalmethods}, both histogramming and RANSAC-based goalpost identification techniques were evaluated in terms of two performance metrics: ``detection rate" and ``distance-by-width". Figure~\ref{fig:results1} presents the detection rate for both algorithms, over a range of distances and coronal body tilt angles. For each correctly identified goalpost, Figure~\ref{fig:results2} presents the calculated distance from goalpost to robot. As the only variable in distance calculation was the calculated pixel width of the goalpost, this metric is an implicit measure of the ``quality" of correctly identifying a post. Both sets of performance results are presented for 6 fixed points on the RoboCup field of known distance, and 3 different sideways body tilt angles; an important consideration due to the oscillations experienced by a biped in motion, which can considerably reduce the performance of traditional detection methods.

\begin{figure}[H]
\begin{center}
\includegraphics[width=7.0cm]{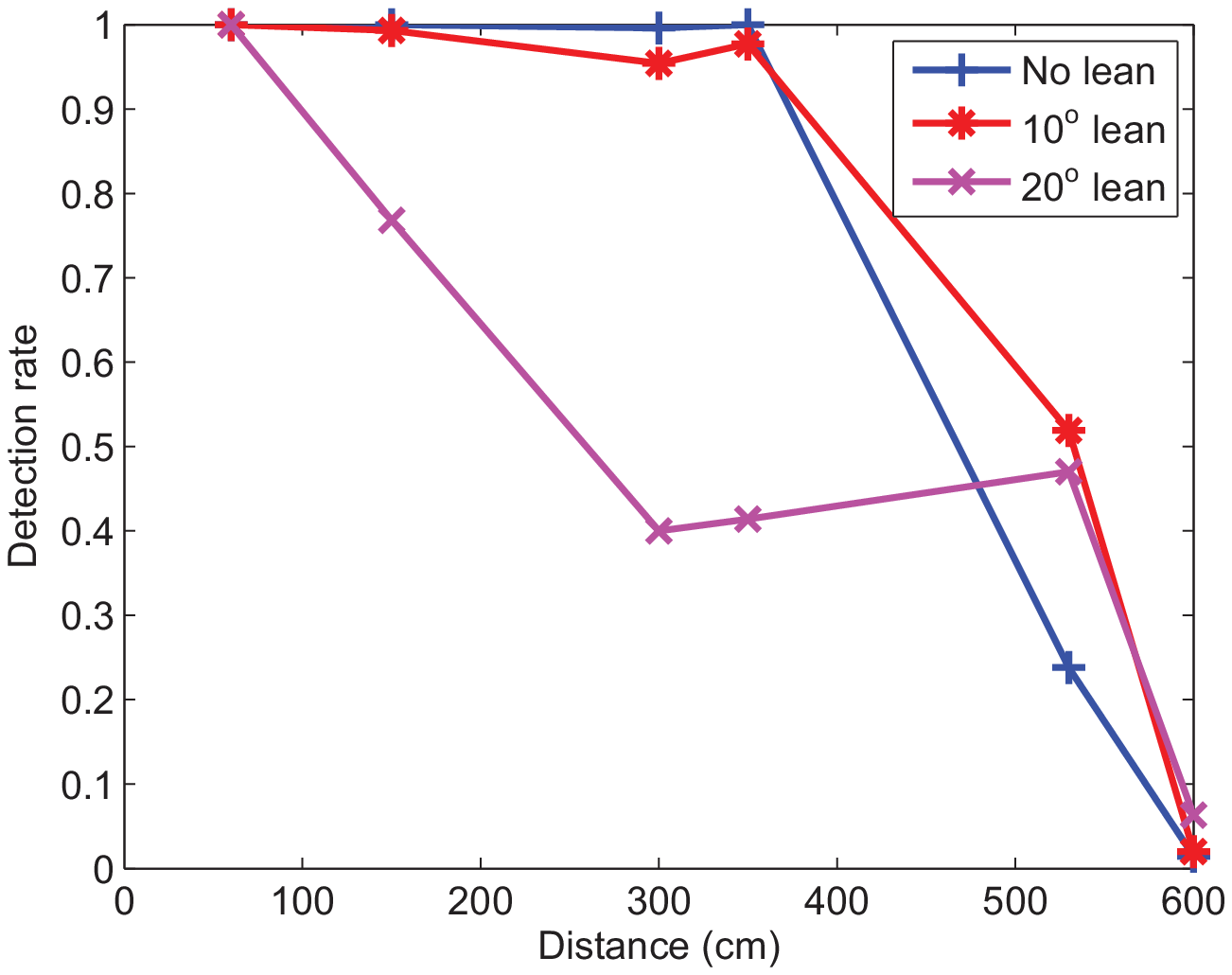}\\ \includegraphics[width=7.0cm]{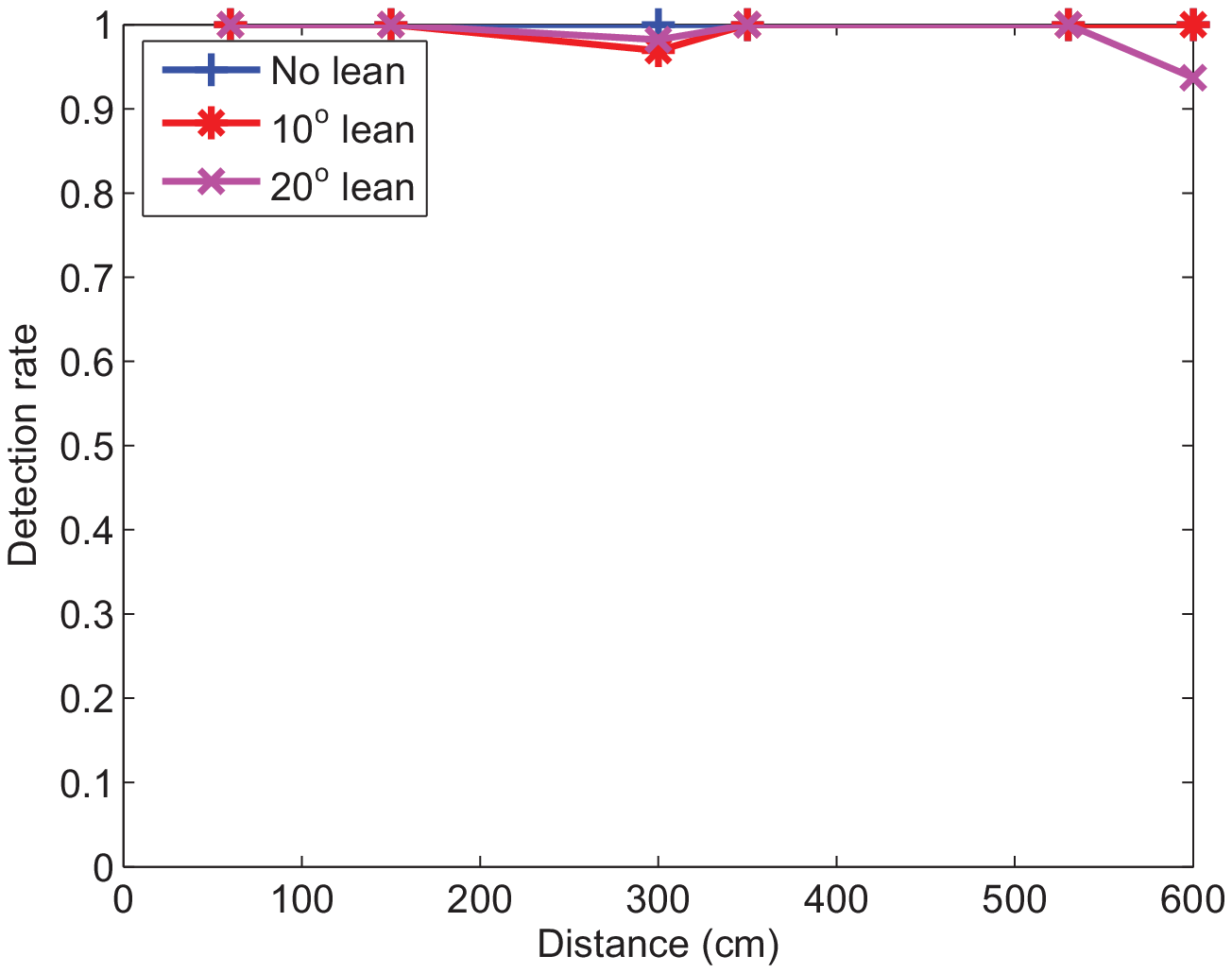}
\end{center}
\caption{Experimental detection rate for histogramming (top) and RANSAC-based (bottom) goalpost identification techniques, as a function of distance from goalpost to robot. A significant performance increase is evident for RANSAC, which is able to accurately identify the majority of goalposts independent of body tilt; an important consideration due to the natural coronal oscillation of a walking bipedal robot. \label{fig:results1}}
\end{figure}

In addition to evaluating the impact of the developed framework on goalpost identification performance, the accuracy of the agent's self-localisation was determined using both histogramming and RANSAC-based techniques. This was accomplished by placing the DARwIn-OP in its initial soccer game position (on the sideline, halfway between the goals and centre circle) with the ball placed on the nearest penalty mark, and having it approach the ball using the goalposts as the only localisation landmarks. This experiment was repeated multiple times, with results demonstrating 18\% and 34\% reductions in positional and angular uncertainty respectively:
\begin{itemize}
\item Histo. pos. uncertainty (cm): $\mu = 31.3, \sigma = 5.5$
\item Histo. ang. uncertainty (rad): $\mu = 0.14, \sigma = 0.06$
\item RANS. pos. uncertainty (cm): $\mu = 25.57, \sigma = 5.1$
\item RANS. ang. uncertainty (rad): $\mu = 0.10, \sigma = 0.05$
\end{itemize}

\begin{figure}[H]
\begin{center}
\includegraphics[width=7.0cm]{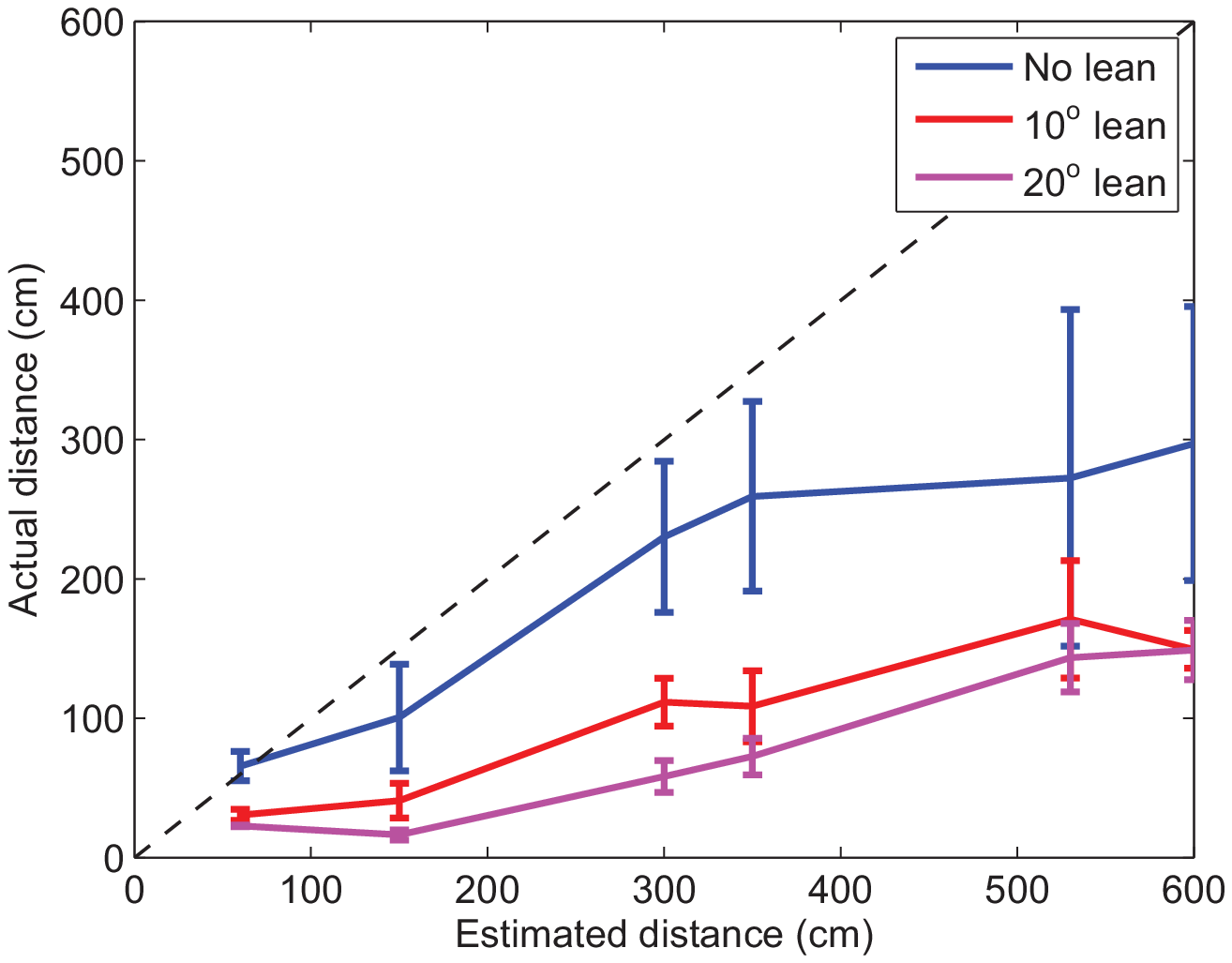}\\ \includegraphics[width=7.0cm]{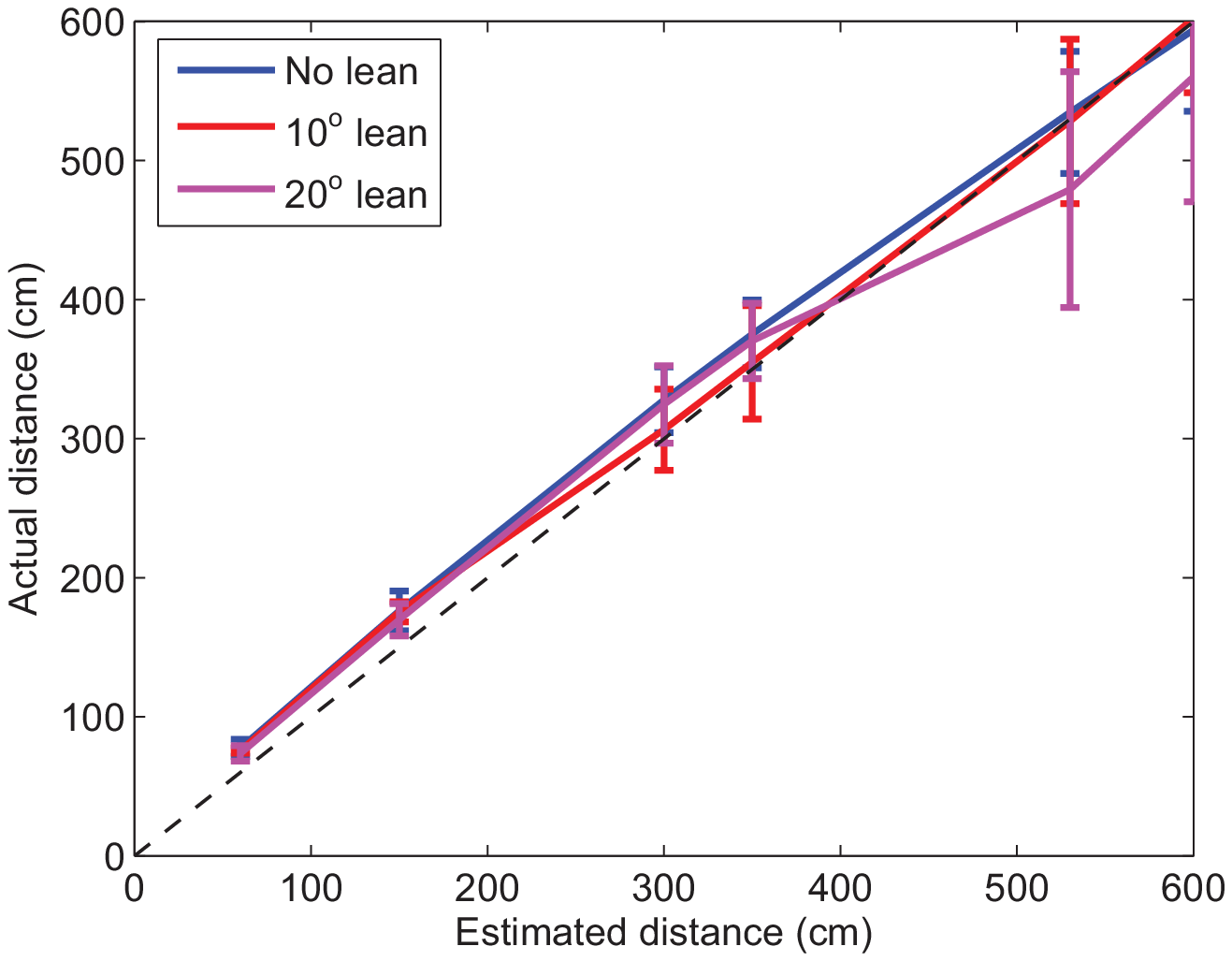}
\end{center}
\caption{Experimental performance for histogramming (top) and RANSAC-based (bottom) goalpost identification techniques. Actual distance from robot to goalposts is plotted against the distance calculated using the ``distance-by-width" method (see Sec. ~\ref{sec:evalmethods}). A significant performance increase is evident for RANSAC, which is able to accurately determine the goalpost distance independent of body tilt. \label{fig:results2}}
\end{figure}

\section{Discussion}

Figure~\ref{fig:results1} demonstrates RANSAC-based goalpost identification as exhibiting a near-perfect detection rate, independent of distance (constrained by RoboCup soccer field length) and body tilt. The detection rate for histogramming is considerably lower, inversely proportional to both tilt and distance from goal (particularly beyond 350 cm). Specifically, the root-mean-square error (RMSE) for histogramming increases from 157.9 to 290.9 cm as the body is tilted to 20 degrees (the maximum tilt observed during normal walking conditions). Conversely, the RMSE for RANSAC is an order of magnitude smaller (19.64 to 30.94 cm) and constant within experimental error.

The observed angular invariance of the RANSAC-based goalpost detection performance follows directly from the framework described in Section~\ref{sec:introduction}. The 2-dimensional projections of goalposts are approximated as rectangular (decomposed into pairs of approximately parallel lines), with candidate points defined as the intersection of the boundaries of colour-segmented regions with a set of equidistant horizontal and vertical “scan-lines”. An extension of the RANSAC algorithm is applied to fit linear models to these candidate points, with pairs of models evaluated against a number of simple rules to ensure they represent the geometry of a goalpost. As no step of this procedure makes assumptions regarding the dominant vertical profile of a goalpost, it is intuitive (and experimentally verified) that the performance of the system will remain constant for any goalpost orientation, subject to experimental noise.

By contrast, traditional 1-dimensional histogramming obtains candidate width by considering the $x$-values of the leftmost and rightmost segments within a peak, and will therefore always result in a candidate representation perpendicular to the ground. When body tilt is present, it follows that the detected width will always exceed actual width, and therefore the estimated distance will always underestimate the actual distance. Due to the natural coronal oscillation of a bipedal gait and the instability of a bipedal kick, it is essential that a goal detection method is able to handle tilt to ensure self-localisation remains accurate during these scenarios.

In a traditional robotics scenario, the agent determines its local-to-global coordinate transformation by inferring its relative position and orientation from observations of salient landmark features. In RoboCup, goalposts are the most commonly considered features for this challenge of self-localisation. It is therefore intuitive that the order of magnitude improvement in goalpost identification resulting from the RANSAC-based technique would correspond with a reduction in the agents positional and angular uncertainty. These improvements were experimentally verified as 18\% and 34\% respectively.


\section{Conclusion}

Reliable self-localisation is critically important in modern robotics, with the accurate identification of salient visual features remaining the most straightforward way of leveraging the performance necessary for industrial applications. Identification of such features by projected geometric profile allows for robustness against spatiotemporal variations in lighting, by removing the dependency on colour classification. This paper has presented a general framework by which the random sample consensus (RANSAC) algorithm can be extended and applied to the identification of such higher-order geometric features, using goalpost identification in humanoid robot soccer as an example application. The framework may be readily extended to generalised robotics and object recognition scenarios by considering colour-independent methods of candidate point generation, such as convolutional filtering for edge detection.

The RANSAC-based method of goalpost identification was demonstrated to yield significant improvements over traditional 1-dimensional histogramming. For a stationary robot, histogramming detection rate was shown to decay to as little as 1\% as distance increases to 6 m (the length of a RoboCup humanoid league field). For goals that are successfully detected, the error in calculated distance was demonstrated to respectively increase from 5 cm to 3 m. Such error makes reliable self-localisation impossible, and worsened further as robot body tilt angle was increased (as experienced during standard walking conditions). In contrast, the proposed RANSAC-based method exhibited near-perfect detection rate invariant of distance or tilt, and was further demonstrated to yield a distance RMSE no greater than 31 cm; an improvement by a full order of magnitude compared to the traditional methodology.

Within RoboCup, it is anticipated that the proposed framework will be readily extended to the identification of more complex geometric features such as the centre circle, goal box and possibly other robots. This improved ability to identify salient features facilitates more accurate self-localisation, therefore allowing for the development of more complex team behaviours and strategies. Applications of higher-order geometric feature identification will also be explored in more generalised robotics and image processing scenarios, including the identification of salient features lacking unique colour-coding.

\section*{Acknowledgement}
NICTA is funded by the Australian Government as represented by the Department of Broadband, Communications and the Digital Economy and the Australian Research Council through the ICT Centre of Excellence program.

\bibliography{BibliographyGoalDetection}
\bibliographystyle{named}

\end{document}